\documentclass{llncs}

\usepackage{tabularx}
\usepackage{array}
\usepackage{paralist}
\usepackage{url}
\usepackage{pbox}
\usepackage{graphicx}
\usepackage[usenames]{color}

\usepackage[numbers,sort,sectionbib]{natbib}

\usepackage{tikz} 

\newcommand{\codiag}{\textit{C-O Diagram}}
\newcommand{\codiags}{\textit{C-O Diagrams}}

\newcolumntype{L}{>{\raggedright\arraybackslash}X}%

\begin{document}

\pagestyle{headings} 

\mainmatter 

\title{Extracting Formal Models from Normative Texts}
\titlerunning{Extracting Formal Models from Normative Texts}

\author{John J. Camilleri\inst{1} \and Normunds Gruzitis\inst{2} \and Gerardo Schneider\inst{1}}
\authorrunning{Camilleri et al.}

\institute{
CSE, Chalmers University of Technology and University of Gothenburg, Sweden
\and
IMCS, University of Latvia\\
\email{\{john.j.camilleri,gerardo\}@cse.gu.se}, \email{normunds.gruzitis@lu.lv}
}

\maketitle

\begin{abstract}
\emph{Normative texts} are documents based on the deontic notions of obligation, permission, and prohibition.
Our goal is model such texts using the \codiag{} formalism, making them amenable to formal analysis,
in particular verifying that a text satisfies properties concerning causality of actions and timing constraints.
We present an experimental, semi-automatic aid to bridge the gap between a normative text and its formal representation.
Our approach uses dependency trees combined with our own rules and heuristics for extracting the relevant components.
The resulting tabular data can then be converted into a \codiag.
\end{abstract}

\section{Introduction}

\emph{Normative texts} are concerned with what must be done, may be done, or should not be done
(\emph{deontic norms}).
This class of documents includes contracts, terms of services and regulations. 
Our aim is to be able to query such documents, by first modelling them in the deontic-based \codiag{}~\cite{MCD+10mvs} formal language.
%
%
Models in this formalism can be automatically converted into networks of timed automata~\cite{Alur1994},
which are amenable to verification. 
There is, however, a large gap between the natural language texts as written by humans, and the formal representation used for automated analysis.
The task of modelling a text is completely manual, requiring a good knowledge of both the domain and the formalism.
In this paper we present a method which helps to bridge this gap, by automatically extracting a partial model using NLP techniques.

We present here our technique for processing normative texts written in natural language and building partial models from them by analysing their syntactic structure and extracting relevant information.
Our method uses dependency structures obtained from a general-purpose statistical parser, namely the Stanford parser~\cite{Klein03}, which are then processed using custom rules and heuristics that we have specified based on a small development corpus in order to produce a table of predicate candidates. 
This can be seen as a specific information extraction task.
While this method may only produce a \emph{partial} model which requires further post-editing by the user, we aim to save the most tedious work so that the user (knowledge engineer) can focus better on formalisation details.




\section{Extracting Predicate Candidates}

The proposed approach is application-specific but domain-independent, assuming that normative texts tend to follow a certain specialised style of natural language, even though there are variations across and within domains.
We do not impose any grammatical or lexical restrictions on the input texts, therefore we first apply the general-purpose Stanford parser acquiring a syntactic dependency tree representation for each sentence.
Provided that the syntactic analysis does not contain significant errors, we then apply a number of interpretation rules and heuristics on top of the dependency structures.
If the extraction is successful, one or more predicate candidates are acquired for each input sentence as shown in Table~\ref{tab:sample-output}.
More than one candidate is extracted in case of explicit or implicit coordination of subjects, verbs, objects or main clauses.
The dependency representation allows for a more straightforward predicate extraction based on syntactic relations, as compared to a phrase-structure representation.

\begin{table}[t]
\small
\caption{Sample input and partial output.}
\label{tab:sample-output}
\begin{tabularx}{\textwidth}{
	>{\hsize=0.48\hsize}L|	
	>{\hsize=0.42\hsize}L|	
   	>{\hsize=0.84\hsize}L|	
   	>{\hsize=1.26\hsize}L|	
	>{\hsize=1.56\hsize}L|	
	>{\hsize=1.44\hsize}L	
}
\cline{1-6}
\multicolumn{1}{l}{\textbf{Refin}.} & \multicolumn{1}{l}{\textbf{Mod}.} & \multicolumn{1}{l}{\textbf{Subject} (S)} & \multicolumn{1}{l}{\textbf{Verb} (V)} & \multicolumn{1}{l}{\textbf{Object} (O)} & \multicolumn{1}{l}{\textbf{Modifiers}} \\
\cline{1-6}
\multicolumn{6}{p{\textwidth}}{\textbf{1.} \textit{You must not, in the use of the Service, violate any laws in your jurisdiction (including but not limited to copyright or trademark laws).}} \\
\cline{1-6}
& F & User & violate & law & V: in User's jurisdiction \newline V: in the use of the Service \\
\cline{1-6}
\multicolumn{6}{p{\textwidth}}{\textbf{2.} \textit{You will not post unauthorised commercial communication (such as spam) on Facebook.}} \\
\cline{1-6}
& F & User & post & unauthorised commercial communication & O: such as spam \newline O: on Facebook \\
\cline{1-6}
\multicolumn{6}{p{\textwidth}}{\textbf{3.} \textit{You will not upload viruses or other malicious code.}} \\
\cline{1-6}
& F & User & upload & virus &  \\
\cline{2-6}
OR & F & User & upload & other malicious code &  \\
\cline{1-6}
\multicolumn{6}{p{\textwidth}}{\textbf{4.} \textit{Your login may only be used by one person - a single login shared by multiple people is not permitted.}} \\
\cline{1-6}
& P & person & use & login of User & S: one \\
\cline{1-6}
\multicolumn{6}{p{\textwidth}}{\textbf{5.} \textit{The renter shall pay all reasonable attorney and other fees, the expenses and costs incurred by owner in protection its rights under this rental agreement and for any action taken owner to collect any amounts due the owner under this rental agreement.}} \\
\cline{1-6}
& O & renter & pay & reasonable attorney & V: under this rental agreement \\
\cline{2-6}
AND & O & renter & pay & other fee & V: under this rental agreement \\
\cline{1-6}
\multicolumn{6}{p{\textwidth}}{\textbf{6.} \textit{The equipment shall be delivered to renter and returned to owner at the renter's risk.}} \\
\cline{1-6}
& O & equipment & [is] delivered [to] & renter & V: at renter's risk\\
\cline{2-6}
AND & O & equipment & [is] returned [to] & owner & V: at renter's risk\\
\cline{1-6}
\end{tabularx}
\end{table}


\subsubsection*{Expected Input and Output}

The basic requirement for pre-processing the input text is that it is split by sentence and that only relevant sentences are included.
In this experiment, we have manually selected the relevant sentences, ignoring (sub)titles, introductory notes etc.
Automatic analysis of the document structure is a separate issue.
We also expect that sentences do not contain grammatical errors that would considerably affect the syntactic analysis and thus the output of our tool.

The output is a table where each row corresponds to a \codiag\ box (clause), containing fields for: 
\begin{inparadesc}
  \item[Subject:] the agent of the clause;
  \item[Verb:] the verbal component of an action;
  \item[Object:] the object component of an action;
  \item[Modality:] obligation (O), permission (P), prohibition (F), or declaration (D) for clauses which only state facts;
  \item[Refinement:] whether a clause should be attached to the preceding clause by conjunction (AND), choice (OR) or sequence (SEQ); 
  \item[Time:] adverbial modifiers indicating temporality;
  \item[Adverbials:] other adverbial phrases that modify the action; 
  \item[Conditions:] phrases indicating conditions on agents, actions or objects; 
  \item[Notes:] other phrases providing additional information (e.g. relative clauses), indicating the head word they attach to.
\end{inparadesc}

Values of the Subject, Verb and Object fields undergo certain normalisation and formatting: head words are lemmatised; Saxon genitives are converted to of-constructions if contextually possible; the preposition ``to'' is explicitly added to indirect objects; prepositions of prepositional objects are included in the Verb field as part of the predicate name, as well as the copula if the predicate is expressed by a participle, adjective or noun; articles are omitted.

A complete document in this format can be converted automatically into a \codiag{} model. 
Our tool however does not necessarily produce a \emph{complete} table, in that fields may be left blank when we cannot determine what to use.
There is also the question of what is considered \emph{correct} output.
It may also be the case that certain clauses can be encoded in multiple ways, and, while all fields may be filled, the user may find it more desirable to change the encoding.

\subsubsection*{Rules}

We make a distinction between rules and heuristics that are applied on top of the Stanford dependencies.
Rules are everything that explicitly follow from the dependency relations and part-of-speech tags. For example, the head of the subject noun phrase (NP) is labelled by \texttt{nsubj}, and the head of the direct object NP---by \texttt{dobj}; fields Subject and Object of the output table can be straightforwardly populated by the respective phrases (as in Table~\ref{tab:sample-output}).


We also count as lexicalised rules cases when the decision can be obviously made by considering both the dependency label and the head word.
For example, modal verbs and other auxiliaries of the main verb are labelled as \texttt{aux} but words like ``may'' and ``must'' clearly indicate the respective modality (P and O).
Auxiliaries can be combined with other modifiers, for example, the modifier ``not'' (\texttt{neg}) which indicates prohibition.
In such cases, the rule is that obligation overrides permission, and prohibition overrides both obligation and permission.

In order to provide concise values for the Subject and Object fields, relative clauses (\texttt{rcmod}), verbal modifiers (\texttt{vmod}) and prepositional modifiers (\texttt{prep}) that modify heads of the subject and object NPs are separated in the Notes field.
Adverbial modifiers (\texttt{advmod}), prepositional modifiers and adverbial clauses (\texttt{advcl}) that modify the main verb are separated, by default, in the Adverbials field.

If the main clause is expressed in the passive voice, and the agent is mentioned (expressed by the preposition ``by''), the resulting predicate is converted to the active voice (as shown by the fourth example in Table~\ref{tab:sample-output}). 


\subsubsection*{Heuristics}

In addition to the obvious extraction rules, we apply a number of heuristic rules based on the development examples and our intuition about the application domains and the language of normative texts.

First of all, auxiliaries are compared and classified against extended lists of keywords.
For example, the modal verb ``can'' most likely indicates permission while ``shall'' and ``will'' indicate obligation.
In addition to auxiliaries, we consider the predicate itself (expressed by a verb, adjective or noun). For example, words like ``responsible'' and ``require'' most likely express obligation.

For prepositional phrases (PP) which are direct dependants of Verb, we first check if they reliably indicate a temporal modifier and thus should be put in the Time field.
The list of such prepositions include ``after'', ``before'', ``during'' etc.
If the preposition is ambiguous, the head of the NP is checked if it bears a meaning of time.
There is a relatively open list of such keywords, including ``day'', ``week'', ``month'' etc.
Due to PP-attachment errors that syntactic parsers often make, if a PP is attached to Object, and it has the above mentioned indicators of a temporal meaning, the phrase is put in the Verb-dependent Time field.

Similarly, we check the markers (\texttt{mark}) of adverbial clauses if they indicate time (``while'', ``when'' etc.) or a condition (e.g. ``if''), as well as values of simple adverbial modifiers, looking for ``always'', ``immediately'', ``before'' etc.
Adverbial modifiers are also checked against a list of irrelevant adverbs used for emphasis (e.g. ``very'') or as gluing words (e.g. ``however'', ``also").

Subject and Object are checked for attributes: if it is modified by a number, the modifier is treated as a condition and is separated in the respective field. 

If there is no direct object in the sentence, or, in the case of the passive voice, no agent expressed by a prepositional phrase (using the preposition ``by''), the first PP governed by Verb is treated as a prepositional object and thus is included in the Object field.

Additionally, anaphoric references by personal pronouns are detected, normalised and tagged (e.g. ``we'', ``our'' and ``us'' are all rewritten as ``$<$we$>$'').
In the case of terms of services, for instance, pronouns ``we'' and ``you'' are often used to refer to the service and the user respectively.
The tool can be customised to do such a simple but effective anaphora resolution (see Table~\ref{tab:sample-output}).

%


\section{Experiments}

In order to test the potential and feasibility of the proposed approach, we have selected four normative texts from three different domains:
\begin{inparaenum}[(1)]
\item PhD regulations from Chalmers University;
\item Rental agreement from RSO, Inc.; 
\item Terms of service for GitHub; and 
\item Terms of service for Facebook. 
\end{inparaenum}
In the development stage, we considered first 10 sentences of each document, based on which the rules and heuristics were defined.
For the evaluation, we used the next 10 sentences of each document.


We use a simple precision-recall metric over the following fields: Subject, Verb, Object and Modality.
The other fields of our table structure are not included in the evaluation criteria as they are intrinsically too unstructured and will always require some post-editing in order to be formalised.
%
%
The local scores for precision and recall are often identical, because a sentence in the original text would correspond to one row (clause) in the table.
This is not the case when unnecessary refinements are added by the tool or, conversely, when co-ordinations in the text are not correctly added as refinements.


\begin{table}[t]
  \centering
  \caption{Evaluation results based on a small set of test sentences (10 per document).}
  \label{tab:results}
  \begin{tabular}{l|ccc|ccc}
      \textbf{Document}
    & \multicolumn{3}{l|}{\bfseries Rules only}
    & \multicolumn{3}{l}{\bfseries Rules \& heuristics} \\
    & \textit{Precision}
    & \textit{Recall}
    & $F_1$
    & \textit{Precision}
    & \textit{Recall}
    & $F_1$ \\
    \hline
    PhD      & 0.66 & 0.73 & 0.69 & 0.82 & 0.90 & 0.86 \\
    Rental   & 0.75 & 0.67 & 0.71 & 0.71 & 0.66 & 0.69 \\
    GitHub   & 0.46 & 0.53 & 0.49 & 0.48 & 0.55 & 0.51 \\
    Facebook & 0.43 & 0.54 & 0.48 & 0.43 & 0.57 & 0.49
  \end{tabular}
\end{table}


The first observation from the results is that the $F_1$ score varies quite a lot between documents; from 0.49 to 0.86.
This is mainly due to the variations in language style present in the documents.
Overall the application of heuristics together with the rules does improve the scores obtained.

On the one hand, many of the sentence patterns which we handle in the heuristics appear only in the development set and not in the test set.
On the other hand, there are few cases which occur relatively frequently among the test examples but are not covered by the development set.
For instance, the introductory part of a sentence, the syntactic main clause, is sometimes pointless for our formalism, and it should be ignored, taking instead the sub-clause as the semantic main clause, e.g. ``User understands that [..]''.

The small corpus size is of course an issue, and we cannot make any strong statements about the coverage of the development and test sets.
Analysing the modal verb \emph{shall} is particularly difficult to get right.
It may either be an indication of an obligation when concerning an action, or it may be used as a prescriptive construct as in \emph{shall be} which is more indicative of a declaration.
%
%
The task of extracting the correct fields from each sentence can be seen as paraphrasing the given sentence into one of the known patterns, which can be handled by rules.
The required paraphrasing, however, is often non-trivial.

\section{Related Work}


Our work can be seen as similar to that of \citet{WynerPeters2011},
who present a system for identifying and extracting rules from legal texts using the Stanford parser and other NLP tools
within the GATE system. 
Their approach is somewhat more general, producing as output an annotated version of the original text.
Ours is a more specific application of such techniques, in that we have a well-defined output format which guided the design of our extraction tool,
which includes in particular the ability to define clauses using refinement.



\citet{Mercatali2005} tackle the automatic translation of textual representations of laws to a formal model, in their case UML.
This underlying formalism is of course different, where they are mainly interested in the hierarchical structure of the documents rather than the norms themselves.
Their method does not use dependency or phrase-structure trees but shallow syntactic chunks. 

\citet{cheng2009} also describe a system for extracting structured information for texts in a specific legal domain.
Their method combines surface-level methods like tagging and named entity recognition (NER) with semantic analysis rules which were hand-crafted for their domain and output data format.

%
%



\section{Conclusion}

Our main goal is to perform formal analyses of normative texts through model checking.
In this paper we have briefly described how we can help to bridge the gap between natural language texts and their formal representations.
%
Though the results reported here are indicative at best (due to the small test corpus), the application of our technique to the case studies we have considered has definitely helped increase the efficiency of their ``encoding'' into \codiags.
%
%
Future plans include extending the heuristics, comparing the use of other parsers, and applying our technique to larger case studies.
%

\subsubsection{Acknowledgements}
This research has been supported by the Swedish Research Council under Grant No. 2012-5746 and partially supported by the Latvian State Research Programme NexIT.

\bibliography{refs}

\end{document}